\documentclass{article}

\usepackage[nonatbib,preprint]{neurips_2023}

\usepackage[utf8]{inputenc} %
\usepackage[T1]{fontenc}    %
\usepackage{hyperref}       %
\usepackage{url}            %
\usepackage{booktabs}       %
\usepackage{amsfonts}       %
\usepackage{nicefrac}       %
\usepackage{microtype}      %

\usepackage{amsmath}
\usepackage{amssymb}
\usepackage{mathtools}
\usepackage{amsthm}
\usepackage{amsfonts}

\theoremstyle{plain}

\theoremstyle{definition}

\theoremstyle{remark}

\usepackage[capitalize,noabbrev]{cleveref}
\usepackage{microtype}
\usepackage{graphicx}
\usepackage{subfigure}
\usepackage{booktabs}
\usepackage[x11names,svgnames,dvipsnames]{xcolor}
\usepackage[most]{tcolorbox}
\usepackage{blindtext}

\title{Evaluating The Robustness of Self-Supervised Representations to 
Background/Foreground Removal}

\author{%
  Xavier F.~Cadet\thanks{Corresponding author: xfc17@ic.ac.uk}\\
  Imperial College London
  \And
  Ranya Aloufi \\
  Imperial College London \\
  \And
  Alain Miranville \\
  Université de Poitiers \\
  \And
  Sara Ahmadi-Abhari \\
  Imperial College London \\
  \And
  Hamed Haddadi \\ 
  Imperial College London \\
}

\begin{document}

\newcommand{\figurecolumnht}[3]{
    \begin{figure*}[ht!]
        \vskip 0.2in
        \begin{center}
            \centerline{\includegraphics[width=\columnwidth]{figures/#1}}
            \caption{#3}
            \label{#2}
        \end{center}
        \vspace{-7mm}
    \end{figure*}
}

\newtcolorbox{somebox}[2][]{enhanced, colback=white, width={\textwidth},
    attach boxed title to top left={yshift={-0.5\baselineskip}, xshift=1cm},
    title={#2},
    boxrule=0.5pt,
    coltitle=black,
    boxed title style={enhanced,
            borderline={0.3mm}{-0.5mm}{LightGreen, solid},
            colframe=white,
            colback=LightGreen,
            colupper={black},
        },
    borderline={0.5mm}{-1mm}{LightGreen,solid},#1%
}

\maketitle

\begin{abstract}
    Despite impressive empirical advances of SSL in solving various tasks, the problem of understanding and characterizing SSL representations learned from input data remains relatively under-explored.
    We provide a comparative analysis of how the representations produced by SSL models differ when masking parts of the input.
    Specifically, we considered state-of-the-art SSL pretrained models, such as DINOv2, MAE, and SwaV, and analyzed changes at the representation levels across 4 Image Classification datasets.
    First, we generate variations of the datasets by applying foreground and background segmentation.
    Then, we conduct statistical analysis using Canonical Correlation Analysis (CCA) and Centered Kernel Alignment (CKA) to evaluate the robustness of the representations learned in SSL models.
    Empirically, we show that not all models lead to representations that separate foreground, background, and complete images.
    Furthermore, we test different masking strategies by occluding the center regions of the images to address cases where foreground and background are difficult. For example, the DTD dataset that focuses on texture rather specific objects.
\end{abstract}

\section{Introduction}

Self-supervised learning (SSL) extracts supervisory signals from the data itself by exploiting its underlying structure~\cite{jingSelfsupervisedVisualFeature2019, chenExploringSimpleSiamese2020, heMomentumContrastUnsupervised2020, balestriero2023cookbook}. Recently, it has been demonstrated that SSL methods such as  SimCLR~\cite{chenSimpleFrameworkContrastive2020}, VICReg~\cite{bardes2022vicreg}, MAE~\cite{he2022masked}, and  DINOv2~\cite{oquab2023dinov2} can achieve superior performance in many downstream vision tasks, including image classification, object detection, and semantic segmentation.
Self-supervised learning methods for visual representation learning rely on the design of a good pretext task~\cite{chenSemiSupervisedUnsupervisedDeep2022}. Examples include autoencoders~\cite{ranzato2007unsupervised, he2022masked}, clustering~\cite{coates2011analysis, caron2018deep}, instance-level discrimination~\cite{wuUnsupervisedFeatureLearning2018, zhaoSelfSupervisedVisualRepresentations2021}, and contrastive learning~\cite{purushwalkamDemystifyingContrastiveSelfSupervised2020}.

Predicting missing parts of the input is one of the standard tasks for SSL pretraining~\cite{zhao2021self, xie2022simmim}.
While this problem has limited the performance improvement from SSL in vision, new SSL techniques, such as MaskCo~\cite{zhao2021self} and SimMIM~\cite{xie2022simmim}, have begun to surpass accuracy records by using masked image modeling approaches to advance the scaling performance of visual models. Such observations provide insights into the inner-workings of SSL learning performance. However, it remains unclear how SSL methods are robust in learning representations where a portion of the input is masked or missing, since most of the advances in SSL were developed using a highly curated dataset, ImageNet~\cite{dengImageNetLargeScaleHierarchical2009, grillBootstrapYourOwn2020, zhangSelfSupervisedVisualRepresentation2020,goyal2021self, caronUnsupervisedLearningVisual2021}.
To evaluate this, we list the fundamental questions we aim to address. We provide tentative answers to these questions through empirical studies.

\vspace{-3mm}
\begin{somebox}{Fundamental Questions}
    \paragraph{Q1.} \textit{%
        How does the performance of self-supervised learning compare to supervised learning when dealing with missing information?}

    \paragraph{Q2.} \textit{How similar are Self-Supervised representations of inputs with missing information?}

    \paragraph{Q3.} \textit{How does the type of missing information affect the performance of self-supervised learning algorithms?}
\end{somebox}

\subsection{Our Contributions}
We consider linear evaluation protocol and similarities metrics to assess robustness of features extracted by Self-Supervised models. We examine how representations are affected by a lack of information in the inputs, based on two main assumptions
(i) Self-Supervised representations should be transferable and (ii) Self-Supervised representations of disjoint views should have low similarities. We provide empirical results based on multiple experiments over various datasets.
\begin{itemize}
    \item First, we evaluate the robustness of representations extracted by SSL models under different masking strategies.
    \item Then, we investigate the similarities of the representations of images views that have no intersections, and show that the State-Of-The-Art DINOv2 model leads to representation similarities matching those from the baseline Supervised model.
    \item Finally, we show that depending on the Self-Supervised training strategies the representations of masked input can be close to one another and separated by the model.
\end{itemize}
\section{Related Work}

\textbf{Self-supervised Learning.}
Self-supervised Learning has shown promising results over the past years, closing the gap with Supervised Learning methods \cite{chenSimpleFrameworkContrastive2020, heMomentumContrastUnsupervised2020, grillBootstrapYourOwn2020}.
According to Balestriero et. al, in~\cite{balestriero2023cookbook} SSL can be classified into four broad families: the Deep Metric Learning family (e.g., SwAV~\cite{caronUnsupervisedLearningVisual2021} and SimCLR~\cite{chenSimpleFrameworkContrastive2020}), the Self-Distillation family (e.g., DINOv2~\cite{oquab2023dinov2}), the Canonical Correlation Analysis family (e.g., VICReg~\cite{bardes2022vicreg}), and the Masked Image Modeling family (e.g., MAE~\cite{he2022masked}).
Self-supervised pretraining is commonly evaluated on image classification~\cite{balestriero2023cookbook}, by training a linear classifier using pretrained representations, but there are other evaluation methods like visual evaluation~\cite{bordes2023surprisingly}.
Bordes et al.~\cite{bordes2023surprisingly} suggest, for example, training a conditional generative diffusion model using an SSL representation.
Thus, one can gain some insight into the representation by analyzing which information remains constant across different generated samples and which information does not remain constant.
In contrast to other approaches, we are inspired by literature~\cite{laakso2000content, kornblith2019similarity} that uses representational similarity analysis to compare the robustness of supervised and SSL models to corrupted data inputs.

\textbf{Masked Inputs and Representation Learning.}
Masked image modeling (MIM) has recently become a popular self-supervised visual pretraining method due to its impressive fine-tuning performance on a variety of downstream computer vision tasks~\cite{doersch2015unsupervised, chen2020generative, trinh2019selfie}.
Beyond representation learning, masked image modeling is a classical computer vision problem, named image inpainting.
This problem has been extensively studied in computer vision for a long time~\cite{pathak2016context}, aiming for improving the inpainting quality and without connecting to self-supervised representation learning.
The iGPT, ViT, and BEiT research teams have recently re-explored this approach with regard to modern vision transformers~\cite{caron2021emerging}, and they demonstrate a strong potential for representation learning by introducing specific designs for some components.
These include clustering pixels, predicting mean color, and tokenizing using a block-wise masking strategy in addition to dVAE networks~\cite{xie2022simmim}.
We explore such techniques as a means of constructing approximate views, based on multiple benchmark datasets, to simulate the corruption that occurs in real-world scenarios.

\textbf{Representation Similarities.}
Recent research has attempted to explain neural networks' behavior by comparing representations between layers and among different trained models.
We build upon previous studies investigating neural network representations' similarity~\cite{laakso2000content} to evaluate the robustness of different self-supervised learning methods when learning from masked or missing data.
In particular, we compare the learning performance of various SSL methods that belong to different families using Canonical Correlation Analysis (CCA)~\cite{hardoon2004canonical} and Centered Kernel Alignment (CKA)~\cite{kornblithSimilarityNeuralNetwork2019}.
CCA is a Multivariate Statistics technique that identifies and measures the relationships between two sets of variables, or to find patterns that represent the variation in the data.
It constructs pairs of linear combinations (called canonical variables) from the original variables that are orthogonal within each set and maximally correlated between sets.
In contrast, CKA analyzes representational similarity matrices and is not invariant under invertible linear transformations~\cite{kornblith2019similarity}.
This enables CKA to identify correspondences between representations extracted from networks trained from different initializations.

\section{Methodology}

\subsection{Preliminaries}
In this study, we investigate how well self-supervised representations perform with incomplete or corrupted data, assuming that these representations should be adaptable and maintain satisfactory performance in real-world scenarios. To test this, we evaluate the similarities between two self-supervised representation vectors using Canonical Correlation Analysis (CCA) and Centered Kernel Alignment (CKA) cross variations of datasets. Let's denote the first representation vector as $X = [x_1, x_2, ..., x_n]$ and the second representation vector as $Y = [y_1, y_2, ..., y_n]$, where $n$ represents the dimensionality of the vectors. We use different similarity metrics to quantify the degree of similarity between $X$ and $Y$. By applying these statistical approaches, we can quantitatively evaluate the similarities between self-supervised representations to provide valuable insights into the alignment and linear relationships between these representations.

\subsection{Segmentation As Missing Information}
Exploring segmentation as missing information in self-supervised learning has the potential to improve the robustness and scalability of representation learning. By inferring missing segmentations from visual cues, it is possible to develop more comprehensive and holistic representations that encompass both global context and local details. Furthermore, this approach could contribute to the advancement of self-supervised learning techniques and contribute to utilizing visual data more efficiently and effectively in a variety of applications, such as object recognition, scene understanding, and image synthesis. We leverage the Segment Anything Model (SAM) \cite{kirillov2023segany} to separate the object associated to the target class in the dataset. For instance, if the class to predict in the image is apple, we obtain the segmentation mask of the apple and generate two variations of the image: the \textbf{F}oreground and the \textbf{Ba}ckground images.
These two variations share no pixels.

\subsection{Region Masking as Missing Information}
By considering region masking as missing information, self-supervised models are encouraged to learn robust and discriminative representations that can effectively handle unseen or partially masked regions in real-world scenarios. By using this approach, the model is better able to capture the underlying semantics and spatial relationships of objects without explicitly annotating the regions. We generate Circular Masks, ensuring that the mask covers the same distance in all direction from the center of the image. From these masks, we generate two other variations of the image: the \textbf{C}enter and \textbf{Bo}rder versions of the image.
The two variations have no pixels in common.

\figurecolumnht{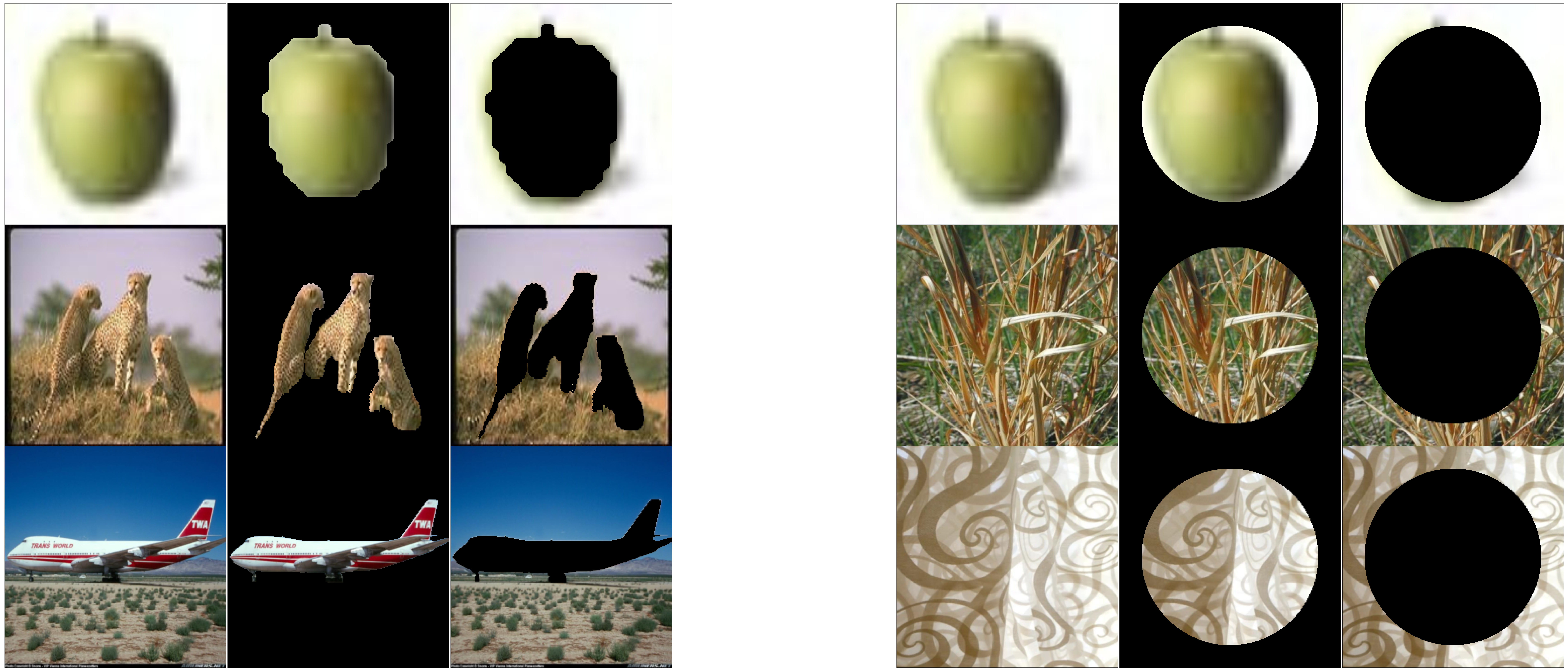}{figures:mask_examples}{Examples of the 5 variations explored. On the left we show the \textbf{O}riginal, \textbf{F}oreground, and \textbf{Ba}ckground from top to bottom, on CIFAR-100, Caltech101, FGVC Aircraft. On the right we show the \textbf{O}riginal, \textbf{C}enter, \textbf{Bo}rder from top to bottom, CIFAR-100, DTD and DTD}
\section{Experimental Setup}
\textbf{Datasets.}
We consider 4 datasets.
We consider the FGVC AirCraft \cite{majiFineGrainedVisualClassification2013}, Caltech-101 \cite{fei-feiLearningGenerativeVisual2004}, CIFAR-100 \cite{krizhevskyLearningMultipleLayers2009}, DTD \cite{cimpoiDescribingTexturesWild2014}.
This selection of dataset covers different sample size, resolution. The DTD dataset provide a case where the objective is focusing on the texture rather than a specific object.

\textbf{Pretrained Models.}
We consider the default Supervised ResNet50 model from the Pytorch \cite{stevensDeepLearningPyTorch2020} library.
Only the Supervised architecture has access to labels during the training phase.
We obtain pretrained weights of ResNet50\cite{heDeepResidualLearning2015} models.
All these ResNet50 models are pretrained on ImageNet-1K.
We obtain the pretrained models, from different source, the majority of the weights come from \cite{goyal2021vissl}.
We selected the best performing  models based on the linear classification benchmark.
Additionally we consider models based on the Vision Transformers (ViT) \cite{dosovitskiy2021an}.

\textbf{Masking Strategies.}
\label{section:setup:masking}
We generate different versions of the datasets by applying different masks.
We consider an Image Masking Scheme with Circular Masks and using Masks obtained from a Segmentation Tool.
By construction, for the Circular Masks the Center and Border images share no pixels. Similarly, the Foreground an Background images share no pixels.

\textbf{Feature Extraction.}
For each pretrained model, we freeze the weights and replace the final layer by an identify function.
For each dataset we resize the images to $224$ pixels along the shorter side followed by center cropping to obtain images of size $224 \times 224$. We then normalize the images using the mean and standard variation obtained on ImageNet-1K.
Finally, we apply the different masking strategies to obtain the 5 variants of the dataset.
The Original images, the Foreground only, Background only, Center only and Border only.

\textbf{Linear Evaluation.}
\label{section:setup:linear_evaluation}
We processed to 5-fold cross validation using the feature extracted using the different models.

\section{Results}
\begin{table*}[t]
\caption{Linear Evaluation on CIFAR-100 and Caltech101, reported results are based on average Balanced Accuracy over 5-folds.}
\label{table:linear_evaluation}
\vskip 0.1in
\begin{center}
\begin{small}
\begin{sc}
\begin{tabular}{cccccc}
\toprule
Dataset & \multicolumn{5}{c}{cifar100} \\
Variation & Original & Foreground & Background & Center & Border \\
\midrule
supervised & $0.59_{(0.0)}$ & $0.54_{(0.01)}$ & $0.28_{(0.01)}$ & $0.52_{(0.0)}$ & $0.33_{(0.0)}$ \\
swav & $0.65_{(0.0)}$ & $0.57_{(0.0)}$ & $0.3_{(0.0)}$ & $0.54_{(0.0)}$ & $0.33_{(0.01)}$ \\
vicreg & $0.62_{(0.01)}$ & $0.57_{(0.0)}$ & $0.28_{(0.01)}$ & $0.54_{(0.0)}$ & $0.33_{(0.0)}$ \\
dinov2\_vitg14 & $0.86_{(0.0)}$ & $0.74_{(0.0)}$ & $0.31_{(0.01)}$ & $0.77_{(0.0)}$ & $0.44_{(0.01)}$ \\
mae & $0.46_{(0.01)}$ & $0.36_{(0.01)}$ & $0.19_{(0.0)}$ & $0.32_{(0.01)}$ & $0.22_{(0.0)}$ \\
\midrule
Dataset & \multicolumn{5}{c}{caltech101} \\
Variation & Original & Foreground & Background & Center & Border \\
\midrule
supervised & $0.93_{(0.01)}$ & $0.89_{(0.01)}$ & $0.75_{(0.01)}$ & $0.9_{(0.0)}$ & $0.62_{(0.01)}$ \\
swav & $0.91_{(0.01)}$ & $0.87_{(0.02)}$ & $0.72_{(0.01)}$ & $0.88_{(0.01)}$ & $0.55_{(0.02)}$ \\
vicreg & $0.92_{(0.01)}$ & $0.88_{(0.01)}$ & $0.73_{(0.01)}$ & $0.89_{(0.01)}$ & $0.56_{(0.02)}$ \\
dinov2\_vitg14 & $0.97_{(0.01)}$ & $0.96_{(0.01)}$ & $0.89_{(0.01)}$ & $0.98_{(0.0)}$ & $0.83_{(0.01)}$ \\
mae & $0.69_{(0.03)}$ & $0.71_{(0.02)}$ & $0.37_{(0.02)}$ & $0.69_{(0.02)}$ & $0.25_{(0.02)}$ \\
\midrule
\end{tabular}
\end{sc}
\end{small}
\end{center}
\end{table*}
In the following we first analyze the linear evaluation performance achieved using features extracted by the different models across dataset variants.
Then, we consider the representations similarities across variants using first CCA, then CKA.
Finally, we discuss the representation proximity using the information from the nearest neighbors.

\figurecolumnht{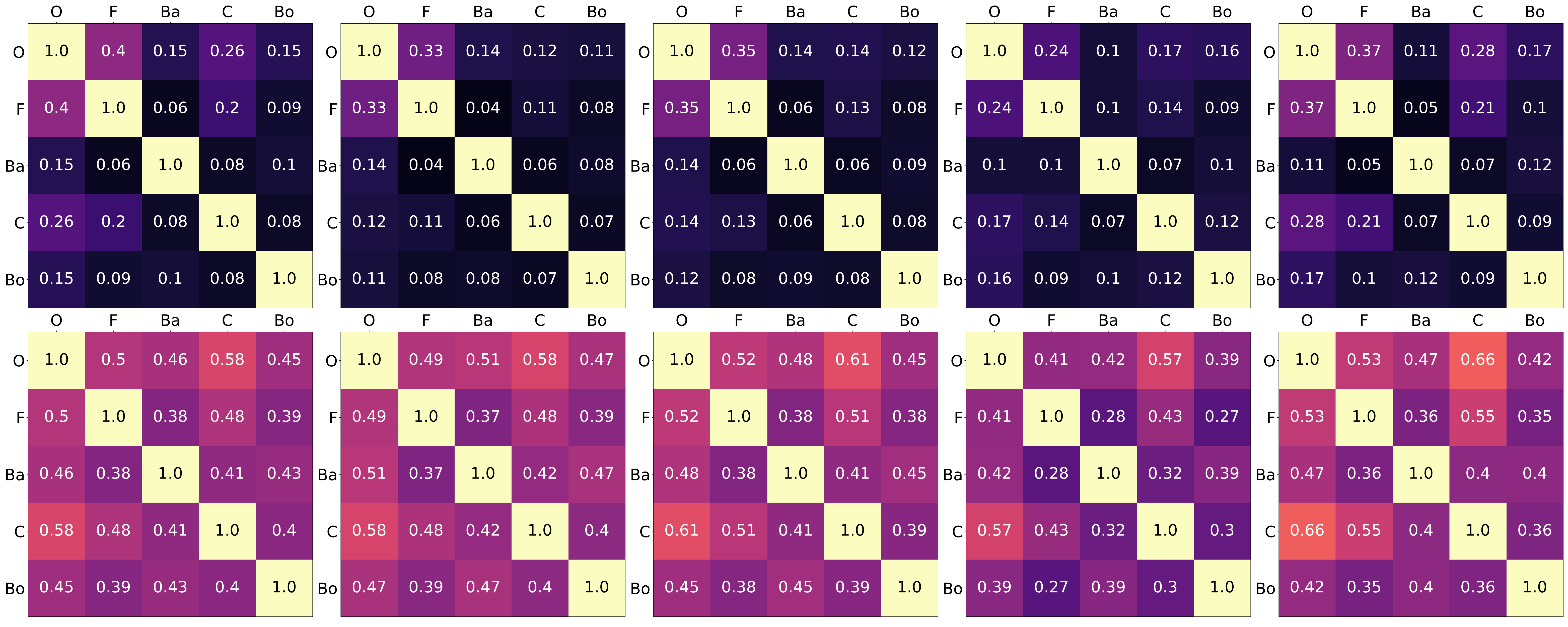}{figure:cca}{Matrices based on the Mean Squared CCA correlation $R^{2}_{CCA}$. The first row is based on CIFAR-100, the second row on Caltech101. From left to right, the matrices are associated with the representation extracted by the Supervised, SwAV, VICReg, MAE and DINOv2 models. \textbf{O}riginal, \textbf{F}oreground, \textbf{Ba}ckground, \textbf{C}enter, \textbf{Bo}rder.}
\figurecolumnht{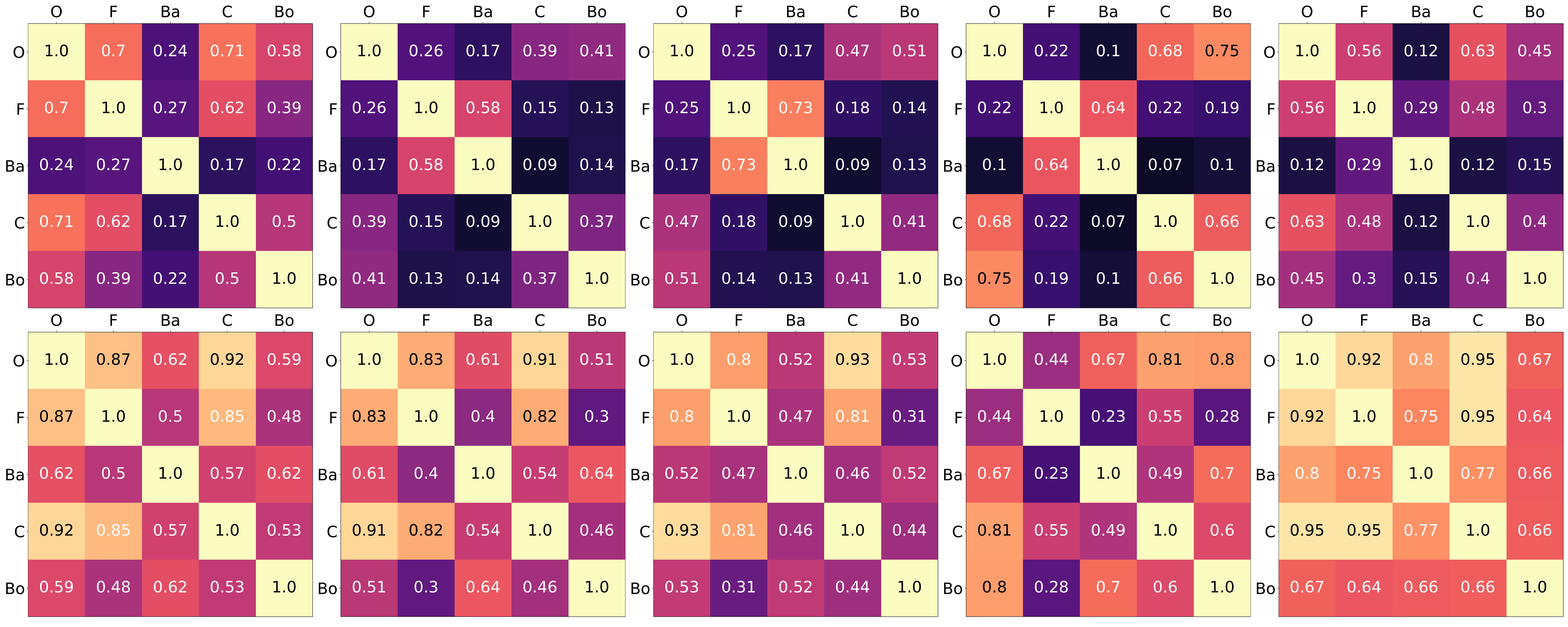}{figure:cka}{Matrices based on the CKA scores. The first row is based on CIFAR-100, the second row on Caltech101. From left to right, the matrices are associated with Supervised, SwAVm VICReg, MAE and DINOv2. \textbf{O}riginal, \textbf{F}oreground, \textbf{Ba}ckground, \textbf{C}enter, \textbf{Bo}rder.}

\figurecolumnht{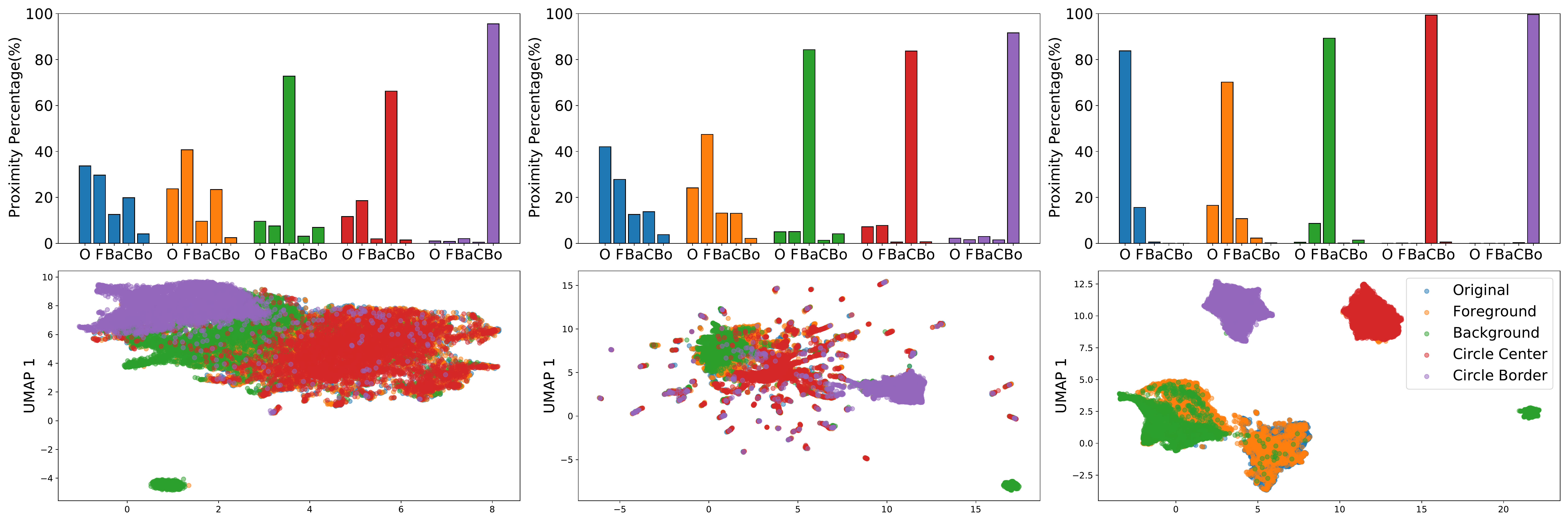}{figure:proximity}{The first row indicates how each variant neighboring points are related to other variants.
    These observations are based on the representations extracted on CIFAR-100. From left to right, we consider the Supervised, DINOv2 and MAE models. The second row depicts the UMAP obtained on $10\%$ of the representations.}

\subsection{Linear Evaluation of the image variations}
We report the results of Linear Evaluation on CIFAR-100 and Caltech-101 in Table \ref{table:linear_evaluation}.
Results on the other datasets can be found in the Supplementary Material.
All the models have consistently higher classification accuracy when using the Original image rather than the Foreground, indicating that there is additional contextual information that is extracted by the models.
This contextual information is then leveraged by the linear classifiers.
DINOv2 obtains better performance when using the Center variant compare to the Foreground variant. For the Foreground variant, there is no guarantee that the entirety of the object will be located in the center of the image, even-though most common datasets locate the object in the center of the image.
For example, in the Caltech-101 dataset, the images which have human faces are target, are not necessarily centered and part of the faces are outside the Center variant.
Nonetheless, all the models except MAE reach higher balanced accuracy on the Original image than on the Foreground, and the Center variant leads to higher performance compared to the Foreground variant. That could indicate that the Self-Supervised models tend to learn additional contextual information from the background.
When evaluated on only the Background version of the dataset, all models except the MAE-based were able to achieve relatively high balanced accuracy, above $70\%$ balanced accuracy.
This could indicate that the dataset lacks background diversity for the different classes and that using only the background information is sufficient on this dataset.

Our findings, highlight that leveraging segmentation to generate different variations of the datasets could help to improve the robustness of the models.

\subsection{Representation Similarity using CCA}
We wonder if the linear evaluation balanced accuracies give enough information about the robustness of the representations learned by the Self-Supervised models.
We report the CCA matrices in figure \ref{figure:cca} for CIFAR-100 and Caltech-101.
From CIFAR-100, we observe that across all models, the representations of the Original images are most similar to the Foreground variants with similarity scores ranging from $0.24$ for the MAE model, to the $0.4$ for the Supervised model.
This trend matches with the observations from the Linear Evaluations.
Furthermore, for both the Supervised and DINOv2 models, the Original and Center variants of the datasets are the most similar followed by the Center to Foreground representations.
On the Caltech101, we observe that across all models the most similar representations are those of the Original and Center images.
This could come from the smaller size of the objects of interest in these datasets with respect to the size of the images.
As such, the Original and Foreground images would lead to less similar representations.
This is further supported by the higher level of similarities between Center and Foreground as well as Border and Background.

We observe that DINOv2 follows similar learning patterns to the Supervised model compared to other Self-supervised models. Nonetheless, the balanced accuracies achieved using the representations extracted by DINOv2 are higher than those obtained using the representations from the Supervised Model.

Our findings call for further research investing whether to design new pretext tasks, or to focus on creating novel preprocessing steps to obtain curated datasets.

\subsection{Representation Similarity using CKA}
We further compare the representations using the CKA, and provide the results in Figure \ref{figure:cka}
We observe that the Supervised and DINOv2 models have higher similarity scores between the Original and Center variants, followed by Original and Foreground.
On CIFAR-100 all the models have lower similarity between the Center and Background representations. Which is to expect since the objects are located in the center of the images.
Furthermore, the MAE model also shows really low similarity between the Background and Original, and Border and Background.
On the Caltech101 dataset, the DINOv2 representations lead to are mostly similar across all variants, with score all above $0.64$.
This also shows that while the CCA indicated that the Center and Background representations were not that similar, the CKA indicates that for DINOv2 the Center and Background representations are highly similar with a score $0.77$.
These corroborate the Linear Evaluation performance observed, with the high balanced using DINOv2's representations on only background images.
Given the pretrained task of the MAE model, it was expected that it would be more robust to the masking strategies.

\subsection{K-Neighbors in the representation space}
We then ask: \textit{Are samples in the representations neighboring samples from the same variants ?}
Through the K-Nearest Neighbors analysis, in Figure \ref{figure:proximity}, we observe that the Supervised and DINOv2 Models leads to Original representations that are close to a mix of others. Indeed, only $40\%$ of the Original and Foreground samples have variant-matching neighbors.
Meanwhile, in the MAE representation space the different variant are mostly matching their neighbors. Indeed, all variants have more than $80\%$ of their closest neighbors belonging to the similar variant.

We further visualize the representation spaces using UMAP.
This highlights that MAE leads to disjointed representation from different variants. Interestingly, it isolated the Center and Border representations from the other variants.

\section{Discussion and Limitations}
Understanding the robustness of self-supervised representations is crucial for assessing their practicality and reliability in real-world applications. By investigating the performance of these representations under varying degrees of data incompleteness or corruption, we can identify potential limitations and design strategies to enhance their robustness. We studied the effect of Foreground / Background removal and input masking on the representations extracted by Self-Supervised models. We investigated how masked input impacted Self-Supervised models compared to Supervised models, and showed that the DINOv2 model leads to representations similarity that match those of the reference supervised model.
Despite these similarities, the learned representation lead to higher balanced accuracies scores.

\textbf{Limitations:}
In this work we only considered segmentation and masking as means of missing information. Further settings and strategies could be explored.
Our study could be extended to other tasks and diverse datasets.

Further analysis, should be conducted to determine whether developing novel pretext tasks, or defining novel preprocessing steps to obtain curated datasets would be most beneficial.

\section*{Broader Impact}
By examining the performance of self-supervised representations under masked or missing inputs, we gain insights into their ability to handle real-world scenarios where data may be corrupted, occluded, or partially available. Our exploration contributes to the continuous development and improvement of self-supervised learning techniques, allowing for the creation of more effective and adaptable models in the face of imperfect or incomplete data.

\begin{ack}
  The authors would like to thank Sandra Siby for numerous comments, technical questions, references, and invaluable suggestions for the presentation that led to an improved text.
  Xavier F. Cadet is supported by UK Research and Innovation [UKRI Centre for Doctoral Training in AI for Healthcare grant number EP/S023283/1].
  This research is supported by UK DRI/MRC Infrastructure for a Trusted Research Environment to Support Cost-effective and Flexible Processing of Healthcare Data grant and UK DRI Translational Machine Intelligence Programme (UKDRI-7002)
  For the purpose of open access, the authors have applied a Creative Commons Attribution (CC BY) license to any Author Accepted Manuscript version arising.
\end{ack}

\bibliographystyle{unsrt}
\bibliography{references}

\end{document}